\documentclass[10pt,twocolumn,letterpaper]{article}

\usepackage{cvpr}      
\usepackage{algorithm}
\usepackage{algorithmic}
\usepackage{multirow}
\definecolor{cvprblue}{rgb}{0.21,0.49,0.74}
\usepackage[pagebackref,breaklinks,colorlinks,allcolors=cvprblue]{hyperref}
\usepackage{algorithm}
\usepackage{algorithmic}
\usepackage{booktabs}   
\usepackage{tabularx}   
\usepackage{bm}
\def\paperID{} 
\def\confName{CVPR}
\def\confYear{2026}

\title{SignReasoner: Compositional Reasoning for Complex Traffic Sign Understanding via Functional Structure Units}

\author{
	\begin{tabular}{c}
	Ruibin Wang \quad Zhenyu Lin \quad Xinhai Zhao \\  
	Huawei Noah's Ark Lab \\ 
	{\tt\small \{wangruibin4, linzhenyu2, zhaoxinhai1\}@huawei.com} 
	\end{tabular}
}

\begin{document}
\maketitle
\begin{abstract}
Accurate semantic understanding of complex traffic signs—including those with intricate layouts, multi-lingual text, and composite symbols—is critical for autonomous driving safety. Current models, both specialized small ones and large Vision Language Models (VLMs), suffer from a significant bottleneck: a lack of {compositional generalization}, leading to failure when encountering novel sign configurations.
To overcome this, we propose {SignReasoner}, a novel paradigm that transforms general VLMs into expert traffic sign reasoners. Our core innovation is {Functional Structure Unit (FSU)}, which shifts from common ``instance-based" modeling to flexible ``function-based" decomposition. By breaking down complex signs into minimal, core functional blocks (e.g., Direction, Notice, Lane), our model learns the underlying ``structural grammar", enabling robust generalization to unseen compositions.
We define this decomposition as the {FSU-Reasoning} task and introduce a two-stage VLM post-training pipeline to maximize performance: {Iterative Caption-FSU Distillation} that enhances the model's accuracy in both FSU-reasoning and caption generation; {FSU-GRPO} that uses Tree Edit Distance (TED) to compute FSU differences as the rewards in GRPO algorithm, boosting reasoning abilities.
Experiments on the newly proposed FSU-Reasoning benchmark, TrafficSignEval, show that SignReasoner achieves new SOTA with remarkable data efficiency and no architectural modification, significantly improving the traffic sign understanding in various VLMs.

\end{abstract}
\begin{figure*}[h]
	\centering
	\includegraphics[scale=0.62]{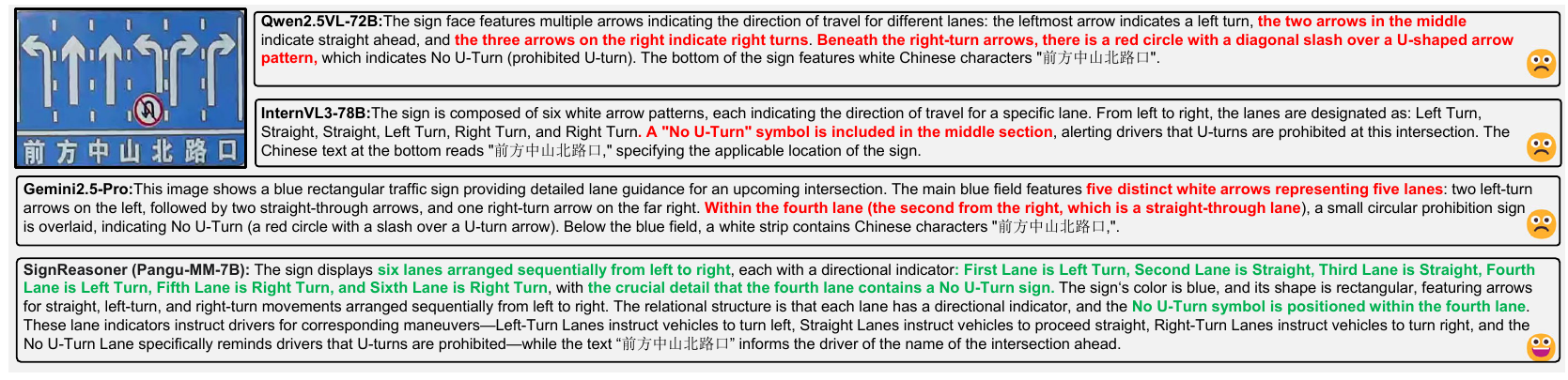}
	\caption{\textbf{Visualizations of complex traffic sign understanding of VLMs.} Existing VLMs, including Qwen2.5-VL-72B \cite{bai2025qwen2}, InternVL3-78B \cite{zhu2025internvl3}, and Gemini 2.5-Pro, demonstrate a significant inability to correctly comprehend traffic signs. Their failures primarily stem from misidentifying the number or type of directional arrows, or failing to accurately associate the U-Turn prohibition symbol with the correct lane. In contrast, our proposed SignReasoner bases on models with only 7B parameters, achieving fully accurate understanding}
	\label{fig0}
\end{figure*}
\section{Introduction}
\label{sect1}
Traffic signs serve as the most direct and codified source of regulatory information in the driving environment. Accurate localization of signs and understanding semantic meaning is essential for an autonomous vehicle to execute safe decision-making and path planning \cite{9046805, 9351818, 000111}. 
Owing to the rapid advancements in robust object detection frameworks (e.g., Faster R-CNN \cite{girshick2015fast, ren2016faster}, YOLO \cite{redmon2016you, cheng2024yolo}, and DETR \cite{zhang2022dino, zhu2020deformable}), traffic sign detection has achieved notable success and high efficacy \cite{zhu2016traffic, wang2023improved, zhang2025tsd}. The research focus is shifting from simply ``seeing" the traffic sign to fully understanding its semantics \cite{guo2024signparser,  chang2025driving, yang2024traffic}. 

Despite the success in classifying the single simple signs, like "stop" or "speed limit 60", fully and accurately reasoning of complex traffic signs—those featuring intricate layouts, multi-lingual text, composite symbols (as illustrated in Fig. \ref{fig0})—remains a critical bottleneck.
We summarize these challenges into two categories. \textbf{1. Arbitrary Open-World Text \& Symbol Recognition}: complex traffic signs incorporate extensive content in various languages (e.g., Chinese, English), numerical information (e.g., street id, distance) and long-tail symbols which mandates models capable of not only precise OCR but robust symbol classification.
\textbf{2. Semantic Reasoning of Text-Symbol Composites}: complex signs have a vast number of the composite integration of multiple symbols and supplementary text (e.g., ``No Trucks" coupled with a ``Ahead 500 Meters" advisory). 
The development of models capable of robustly and efficiently generalizing compositional relationships among constituent elements remains a significant open research challenge.

To tackle these difficulties, current research efforts can be broadly categorized into two major approaches: small specialized models and Large VLMs. The first category decomposes the complex traffic sign understanding tasks (including component detection, content reasoning, and semantic description) and design multi-module networks tailored to solve each sub-tasks. Although promising progresses have been made \cite{guo2024signparser, yang2024traffic}, small models intrinsically lack reasoning and generalization capabilities and are still prone to hard failures when encountering unfamiliar symbol and compositions.
Recently, the academic community have witnessed the rise of VLM in powerful reasoning and generalization capabilities, which, along with their rich pre-training traffic knowledge, makes them suitable tools for the complex sign understanding. However, there have currently been only a very small number of attempts utilizing VLMs for complex traffic sign reasoning, which either simply treats the task as a form of VLM's captioning \cite{yang2025signeye} or just borrows the VLM's encoder structure for traditional multi-tasks processing \cite{chang2025driving}. A method tailored for traffic sign comprehension and readily applicable to current VLMs has not yet been introduced.

In this paper, we introduce SignReasoner, the first framework designed to convert existing VLMs into experts for complex traffic sign understanding without any alteration to VLM's architecture. SignReasoner achieves this by repurposing and orchestrating VLM's inherent capabilities—specifically its symbol recognition, OCR, and reasoning abilities—through a novel traffic sign modeling method and a highly data-efficient training pipeline.

Specifically, we first propose the concept of Functional Structure Unit (FSU) for traffic sign modeling. FSU decomposes a sign into its minimal, core functional building blocks, which fundamentally shift from modeling a sign from an instance-based manner to a flexible function-based way. 
Four essential functional types are defined: Direction, Notice, Lane, and Construction, with each function incorporating specific attributes and stored in a VLM-friendly dictionary format. 
Consequently, any complex traffic sign can be naturally decomposed into a combination of the four basic FSUs. The complex traffic sign understanding task can thus be converted to the task of reasoning to identify FSUs, which we define as FSU-Reasoning. Next, to repurpose the inherent capabilities of VLM to maximize FSU-Reasoning performance, a two-stage post-training pipeline is adopted. 
In the first stage, we propose the Caption-FSU Iterative Distillation, which constructs the SFT samples in a ``Caption-FSU” format and iteratively refine the caption through distillation, converging to optimal FSU-Reasoning performance.
The second stage adopts the GRPO algorithm \cite{shao2024deepseekmath} to boost VLMs reasoning ability on hard samples. A unique Tree Edited Distance (TED) is adopted as the rule-based rewards in GRPO to hierarchically measure FSU differences of ground truth and prediction.

Experimentally, we established the first FSU-reasoning benchmark, TrafficSignEval, to rigorously evaluate VLM's comprehension of various traffic sign contents. Zero-shot evaluation on open-source VLMs, including Qwen \cite{bai2025qwen2} and InternVL \cite{zhu2025internvl3} revealed existing model deficiencies. Our framework, SignReasoner, using only 726 training samples and without any architectural modifications, significantly boosting the FSU-Reasoning capabilities of various VLMs. Ablation studies validate the effectiveness of the proposed FSU traffic sign modeling and the two-stage post-training. In summary, our contributions are:

\begin{itemize}
    \item We propose SignReasoner, a novel traffic sign modeling paradigm centered on the FSU concept, and develop a two-stage post-training process featuring a novel Iterative Caption-Structure Distillation and FSU-GRPO to maximize FSU-Reasoning performance.
    \item We establish the TrafficSignEval benchmark to rigorously evaluate VLMs on complex traffic sign understanding. Experiments confirm that SignReasoner significantly boosts various VLMs for complex sign understanding.
    \item Extensive ablations are conducted to verify each design in SignReasoner. Visualization is provided to more clearly demonstrate the improvement brought by SignReasoner.
\end{itemize}

\section{Related Work}
\label{sec:intro}
\subsection{Complex Traffic Sign Reasoning}
Current research efforts for complex traffic sign reasoning can be broadly categorized into two major approaches: small specialized models and large VLMs. 

\textbf{Small Specialized Models.}
This category focuses on designing lightweight and highly efficient multi-module network architectures. For example, \cite{guo2021learning, guo2024signparser} introduce a framework, integrating component detection, content reasoning, and semantic description generation and achieves remarkable performance through joint optimization. \cite{guo2023visual} extends the framework so as to be applied to traffic scene images. \cite{yang2024traffic} further improves the symbol recognition within different traffic panels and interprets the sign as natural language. However, these small models inherently lacking generalization and reasoning,  rely on exhaustive datasets and suffer failures when facing novel or non-standard sign compositions.
	
\textbf{Large Vision-Language Models.} Recently, VLMs \cite{wang2024qwen2, chen2024internvl, huang2024vlm, pan2025medvlm} have demonstrated formidable generalization and inherent reasoning capabilities in various domains. However, reasoning complex traffic signs using VLMs has seen only limited exploration. \cite{yang2025signeye} treats the complex sign reasoning as a form of visual captioning and construct SFT data in a instance-based caption templates. However, this captioning approach results in unstructured natural language output with hallucination. \cite{chang2025driving} borrows the vision and language encoders from VLM and combine other designed modules to separately perform OCR, elements clustering, content understanding, which is essentially a multi-stage methodology, sacrificing the VLM's inherent end-to-end reasoning by introducing cumulative errors in each stage. 

We also employ VLMs for complex traffic sign reasoning. Differently, our method designs a more generalized modeling approach that simultaneously supports both unstructured and structured outputs. More critically, our specially designed post-training scheme enables end-to-end, accurate and generalized sign reasoning without any modifications to the base model architecture.

\subsection{Post-Training on VLM}
Post-training is a crucial stage for tailoring generic pre-trained VLMs to achieve superior performance in specific downstream tasks. While various post-training methods exist \cite{kumar2025llm}, we primarily focus on those relevant to our paper: CoT Fine-Tune, Distillation and GRPO Training.

\textbf{CoT Fine-Tune.} 
CoT fine-tuning utilizes supervised reasoning annotations to train models to output step-by-step reasoning traces instead of just final answers. 
This process fundamentally improves both interpretability and accuracy on complex tasks and has been widely adapted from prompting \cite{wei2022chain,wang2022self} to full fine-tuning across language \cite{bai2023qwen,touvron2023llama,guo2025deepseek} and multi-modal domains \cite{thawakar2025llamav, xu2025llava}. Our methods firstly introduce this CoT fine-tune into the complex traffic sign understanding and propose the ``Caption-FSU" output format, aiming to help the model achieve more accurate comprehension.

\textbf{Distillation.} This technique transfers the refined capabilities of the main model to smaller architectures. DeepSeek-R1 \cite{guo2025deepseek}, serves as a teacher to smaller architectures (e.g., Qwen \cite{bai2023qwen} or Llama \cite{touvron2023llama}), allowing the smaller models to inherit advanced reasoning capabilities. 
Unlike these methods, we focus on distilling the model's captioning ability specifically within the traffic sign domain. We propose a novel algorithm that distills the same models from different optimization stages and improve this targeted ability to an optimum with acceptable distillation steps.

\textbf{GRPO Training.} Following DeepSeek-R1's successful use of GRPO \cite{shao2024deepseekmath} to significantly boost model reasoning, an increasing number of researchers are adopting this rule-based reward algorithm, focusing on optimizing the objective function \cite{yu2025dapo, liu2025understanding} and extend it to the multi-modal domain \cite{huang2025vision, yang2025r1, meng2025mm}.
Besides, various rule-based rewards are defined under GRPO, successfully improving the downstream perception tasks, such as grounding \cite{liu2025visual,huang2025vision} and segmentation \cite{liu2025seg}. 
In this work, we design a novel TED based
reward for measuring FSU differences and combine with GRPO algorithm to boost the reasoning and generalization ability in the Caption and FSU outputs.

\begin{figure*}[h]
	\centering
	\includegraphics[scale=0.8]{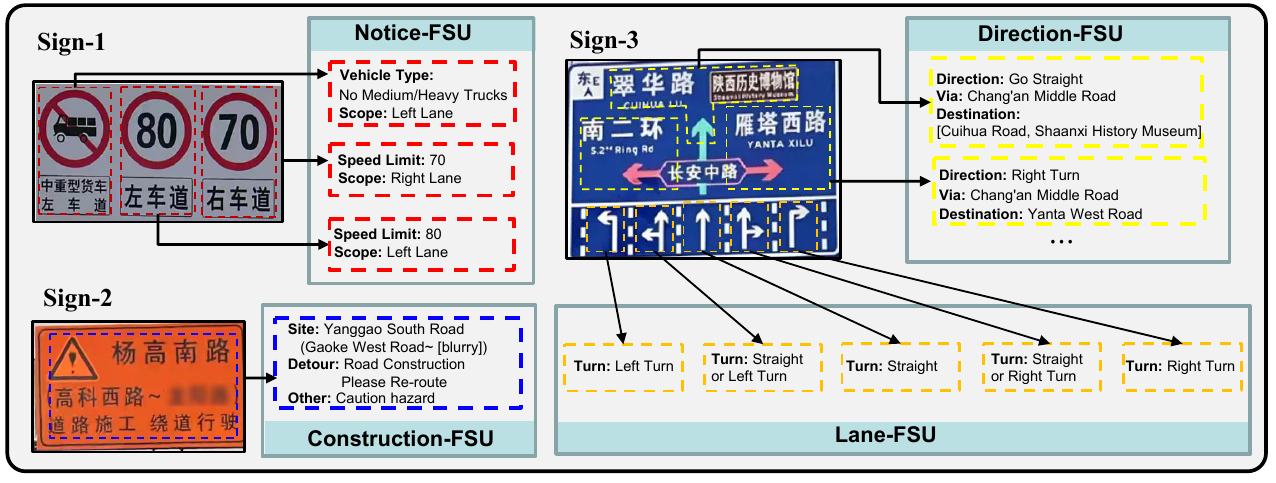}
	\caption{\textbf{Illustration of Functional Structure Units (FSUs) for modeling traffic signs.} Four kinds of FSUs are shown, including Notice, Direction, Construction and Lane. Notably, a traffic sign can either be decomposed into multiple FSUs with same or different functions.}
	\label{fig1}
\end{figure*}

\section{SignReasoner}
\label{sect3}
SignReasoner aims to enhance the existing VLMs in interpretation of complex traffic signs. 
Our approach focuses on two key innovations. 
Firstly, we introduce a novel modeling approach centered on the Functional Structure Unit (FSU), which fundamentally improves the generalization of traffic signs modeling, notably, gives rise to the FSU-Reasoning task for achieving VLM end-to-end traffic sign understanding (Sect.\ref{sect3.1}). 
Secondly, we devise a two-stage training pipeline including  Iterative Caption-FSU Distillation (Sect.\ref{sect3.2}), which systematically fine-tunes VLMs for collaborative enhancement of captioning and FSU-Reasoning, and FSU-GRPO Training (Sect.\ref{sect3.3}) that boosts the reasoning and generalization in hard traffic signs.

\subsection{Functional Structural Unit}
\label{sect3.1}
For the first time, we propose the concept of Functional Structural Unit, a function-based traffic sign modeling approach. FSU is fundamentally a highly formatted dictionary that can be represented as:
\begin{equation}
     \text{FSU} = \{(k_j:v_j)\}_{j=1}^{N}
\end{equation}
where $(k_j:v_j)$ refer to the $j$-th attribute of the FSU represented in a key-value pair, and $N$ is the total number of key-value pairs in one FSU. We provide several examples, as illustrated in Fig. \ref{fig1} for better understanding the FSU.

\textbf{Function-Based Modeling Logic.}  Unlike the Instance-Based logic that focus on modeling specific sign, we define four kinds of functions that frequently appear in a traffic sign and model the corresponding region as one FSU. 
Four kinds of functions are defined, including \textit{Direction} that specifies route guidance and destination information, \textit{Lane} that delineates road organization, turn restrictions, and lane-specific instructions, \textit{Notice} that conveys general regulatory, warning, or informational text and symbols, and \textit{Construction} that alerts drivers to temporary road work, hazards, or diversions. The four categories are designed to be highly orthogonal in function and can be viewed in Fig. \ref{fig1}.

\begin{figure}
	\centering
	\includegraphics[scale=0.67]{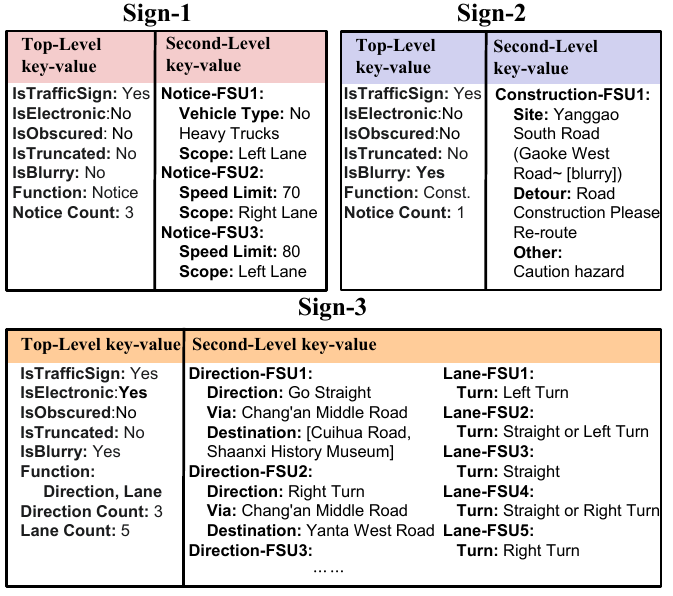}
	\caption{\textbf{Illustration of the Hierarchical Key-Value Schema}. The ``Sign-1,2,3" tags correspond to images in Fig. \ref{fig1} for reference.}
	\label{fig2}
\end{figure}

\textbf{Hierarchical Key-Value Schema.} 
Based on the basic building block of FSU, we introduce a novel Hierarchical Key Schema which decompose a traffic sign into multiple levels for more fine-grained expression:
\begin{equation}
A = \left\{ (k_g:v_g) \right\}_{g=1}^{G} \cup \left\{ (\text{FSU}_i: \left\{ (k_{ij}:v_{ij}) \right\}_{j=1}^{N} ) \right\}_{i=1}^{M}, 
\label{eq2}
\end{equation}
where $A$ refers to the FSU decomposition of a traffic sign. 
It contains the top level key-value pairs $\left\{(k_g:v_g)\right\}_{g=1}^{G}$ that capture $G$ global visual attributes (e.g., Obscured, Blurring, Truncation, etc.) and the second key-value pairs $(\text{FSU}_i: \left\{ (k_{ij}:v_{ij}) \right\}_{j=1}^{N})_{i=1}^{M}$ that detail the information in $M$ FSUs. Essentially, Eq.\ref{eq2} defines the decomposition of a single traffic sign image into a set of global attributes and a collection of $M$ FSUs. A illustration of these top and second levels can be referred to Fig.\ref{fig2}.

\textbf{Hybrid Value Semantics.} The FSU schema adopts a hybrid strategy for populating key values. This approach classifies value types based on their content nature: closed-set values are applied to the binary global attributes (e.g., ``IsBlurry", ``IsObscured") and enumerable function-specific elements (e.g., ``Turn" in Lane-FSU), . Conversely, open-set values are reserved for adaptive content (such as ``Destination" in Direction-FSU) that must be extracted directly from the sign. 
This hybrid strategy effectively frames the closed-set values as straightforward classification and simplifies the open-set component to specialized OCR, mitigating overall modeling complexity.

\textbf{FSU-Reasoning.} 
Based on the proposed FSU, we introduce FSU-Reasoning, a novel, end-to-end task for traffic sign modeling. Given a single sign image $I$ and corresponding prompt $P_{\text{reason}}$, FSU-Reasoning requires the VLMs $M$ to generate correct FSU decomposition (Eq. \ref{eq2}) as:
\begin{equation}
	A \leftarrow M(I, P_{\text{reason}}).
\end{equation}
Essentially, VLM should perform comprehensive reasoning to predict the key-value pairs defined in the FSU. 
FSU-Reasoning is a novel VLM task that firstly frame the complex traffic sign understanding as an straightforward end-to-end task, eliminating the conventional multi-round pipeline (e.g., symbol detection, text recognition, relationship reasoning, etc).
The task necessitates beyond the traffic domain knowledge but general vision-language capabilities including element recognition and counting, OCR, layout analysis, semantic reasoning.
Consequently, FSU-Reasoning serves as an ideal task for evaluating model performance in both general vision and autonomous driving domains. 
We release a benchmark named TrafficSignEval and associated evaluation codes. The detailed information about the benchmark and evaluation protocol can be referred in Sect.\ref{sect4.1}. 

Under TrafficSignEval, we test the zero-shot performance of existing VLMs for FSU-Reasoning. 
As results shown in Tab. \ref{tab1}, the tested VLMs generally has unsatisfactory performance, whether prompted to generate understanding captions or FSU decomposition.
To this end, we introduce a novel two-stage training pipeline which consists of a  Iterative Caption-FSU Distillation and a FSU-GRPO Training, to improve VLM's FSU-Reasoning capabilities.

\subsection{Iterative Caption-FSU Distillation}
\label{sect3.2}
This stage utilizes a ``Caption-FSU" format to organize the initial SFT samples, and designs an Iterative Distillation algorithm to continually refine and update the caption in the SFT samples. Ultimately, this algorithm yields samples with the highest caption quality, training VLMs to excel in both captioning and FSU-Reasoning.

\textbf{Caption-FSU Format.} Inspired by the Chain-of-Thought (CoT) technique \cite{wei2022chain}, which significantly enhances complex reasoning by requiring models to generate intermediate steps before the final answer, we organize our training data in the ``Caption-FSU" format. Namely, before generating the FSU answer, we enforce the model to describe the layout and elements of the traffic sign within a caption. The caption can be viewed as the think process, guiding model generating the correct FSU decomposition. Formally, the SFT data $D$ is organized as:
\begin{equation}
\begin{aligned}
	D &= [C\mid A] \\
	  &= \textless\text{caption}\textgreater{C}\textless/\text{caption}\textgreater\textless\text{FSU}\textgreater{A}\textless/\text{FSU}\textgreater,
	  \label{eq4}
\end{aligned}
\end{equation}
where $C$ refers to sign captions which are concatenated with FSU decomposition (Eq.\ref{eq2}) via the special separator.
In practice, to reuse the VLM's captioning ability, we leverage their originally generated captions without any modifications as $C$. 
By captioning before FSU-Reasoning, we essentially make a tight connection between these two tasks, which forces the model learning better captioning ability so as to derive the more correct FSU outputs. Eventually, the caption-FSU data, which labeled only on FSUs, brings the models simultaneous improvements in captioning and FSU-Reasoning. We term this phenomenon the Synergistic Effect, and leverage it to propose the Iterative Distillation algorithm that further enhances model performance.

\textbf{Iterative Distillation.} The core idea of iterative distillation is straightforward: since training with the ``Caption-FSU" data leads to an improved captioning ability, and high-quality captions result in better FSU decomposition, we iterate this process until reaching the optimal performance.
The complete algorithm is detailed in Alg. \ref{alg1}, where $C^{(t)}$ refers to the caption distilled in $t$ iteration, and $D^{(t)}$ is the updated SFT data, which would be used to train the base VLM $M_{\theta}$, generating the caption-enhanced VLM $M_{\theta^{(t+1)}}$ for next round distillation. Details about the captioning and reasoning prompts are provided in the supplementary.

\begin{algorithm}[t]
\caption{
 Iterative Caption-FSU Distillation
}
\label{alg1}
\begin{algorithmic}[1]
\REQUIRE Traffic sign image $I$, FSU-based annotations $A$ and model parameters from $t$-th iteration $M_{\theta^{(t)}}$, number of iterations $T$, prompts for captioning $P_{\text{caption}}$ and FSU-Reasoning $P_{\text{reason}}$
\FOR{$t = 0$ to $T$}
\STATE Distill the caption $C^{(t)}$ from model $M_{\theta^{(t)}}$:
\STATE $\ \ \ \ \ \ \ \ C^{(t)} \leftarrow M_{\theta^{(t)}}(I, P_{\text{caption}})$.
\STATE Update caption-FSU training data:
\STATE $\ \ \ \ \ \ \ \ D^{(t)} \leftarrow [C^{(t)}|A]$.
\STATE SFT the model parameters via training data $D$:
\STATE $\ \ \ \ \ \ \ \ M_{\theta^{(t+1)}} \leftarrow SFT(M_{\theta}, D, P_{\text{reason}})$.
\ENDFOR

Output $M_{\theta^{(T)}}$
\end{algorithmic}
\end{algorithm}

\begin{figure*}
	\centering
	\includegraphics[scale=0.64]{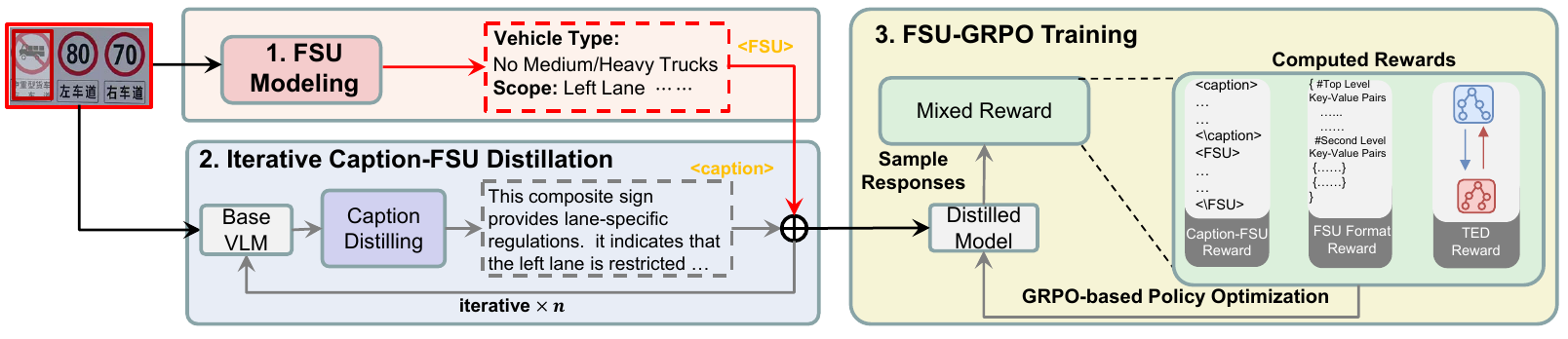}
	\caption{\textbf{Illustration of the SignReasoner overall pipeline.} The FSU Modeling (Sect. \ref{sect3.1}) decomposes the traffic sign multiple into FSUs, which would be concatenated with captions generated from base VLMs for  Iterative Caption-FSU Distillation (Sect. \ref{sect3.2}). Finally, the distilled models would go through the FSU-GRPO Training (Sect. \ref{sect3.3}) to further boost reasoning capacity.}
	\label{fig3}
\end{figure*}

\subsection{FSU-GRPO Training}
\label{sect3.3}
Recent research \cite{guo2025deepseek, liu2025understanding} at the intersection of RL and VLMs has yielded significant conclusions regarding the effectiveness of GRPO \cite{shao2024deepseekmath} and rule-based rewards in facilitating complex reasoning. Following these established findings, we propose the FSU-GRPO training to further boost SignReasoner reasoning ability to solve hard samples. We reuse the GRPO algorithm and focus on designing mixed rule-based rewards. Three kinds of rewards are used.

\textbf{Caption-FSU Reward.} Similar to the general think-answer reward, the Caption-FSU reward $R_\text{C-FSU}$ encourages model outputting in the caption-FSU format as in Eq.\ref{eq4}. It is set to 1 if the output format follows, and 0 otherwise.

\textbf{FSU Format Reward.} This reward $R_\text{FSU}$ encourages the FSU being a parsable dictionary as in Eq.\ref{eq2}. It is set to 1 if the output is parsable, and 0 otherwise.

\textbf{Tree Edited Distance Reward.} To quantify the difference between the predicted and ground truth FSU decomposition we define the tree edited distance reward $R_\text{TED}$ as:
\begin{equation}
\begin{aligned}
	R_\text{TED} = f_\text{Act}(\text{TED}(T_\text{Pred}, T_\text{GT})), \\
	f_\text{Act}(x) = 1 - ({\sigma_1} \cdot \tanh(\frac{x}{\sigma_2}) + \sigma_3),
	\label{eq5}
\end{aligned}
\end{equation}
where $T_\text{Pred}$ and $T_\text{GT}$ refer to trees converted from FSU dictionaries. $\text{TED}(\cdot)$ is the function computing tree edited distance \cite{zhang1989simple, pawlik2015efficient}. $f_{\text{Act}}(\cdot)$ refers to the activation function that projects the distance into [0, 1]. $\sigma_{1,2,3}$ are hyper-parameters re-weight the TED to precisely measure the difference.  

Comparing with string similarity, TED offers two key advantages. Firstly, it more precisely measures the differences across the FSU's hierarchy (i.e., the top- and second-level nodes). Secondly, TED enables a flexible permutation. For the Direction, Notice and Construction FSUs where no specific order exist, we use Hungary Algorithm for nodes matching, thus computing TED in an permutation-invariant manner.
As for the Lane FSUs which should be organized in a left-to-right order, we employ ordered tree structure to compute the TED in a permutation-variant way.

\textbf{Mixed Reward.} Finally, since the TED can only be meaningful when the output follows the ``caption-FSU" format and is parsable, (i.e., $R_\text{C-FSU}=1$, $R_\text{FSU}=1$), we multiply the three individual rewards to obtain the final Mixed Reward $R_\text{Mixed}$, as defined below:
\begin{equation}
	R_\text{Mixed} = R_\text{C-FSU} \cdot R_\text{FSU} \cdot R_\text{TED}
	\label{eq6}
\end{equation} 
The overall pipeline of SignReasoner is illustrated in Fig. \ref{fig3}.

\section{Experiments}
\subsection{Experiment Setup}
\label{sect4.1}
\textbf{Benchmarks.} To evaluate VLMs for FSU-Reasoning, we have newly established TrafficSignEval benchmark, which comprises a total of 195 complex signs encountered in real-world scenarios . 
Different from previous datasets \cite{yang2015towards, stallkamp2012man, zhu2016traffic, guo2024signparser}, TrafficSignEval presents significant composite reasoning difficulty for VLM, covering four categories of Notice, Construction, Direction, and Lane. 
Datasets are manually annotated into FSUs decomposition. Details of the datasets are provided in our supplements. 

\textbf{Models.}
We employ and compare our approach against several open-source VLMs, including Qwen-VL-7B/72B \cite{bai2025qwen2}, InternVL-8B/78B \cite{zhu2025internvl3}, and Pangu-MM-7B. We report the zero-shot prompting capabilities of these models. Moreover, we adopt the SignReasoner pipeline under Qwen-VL-7B and Pangu-MM-7B, yielding the enhanced traffic sign understanding models SignReasoner (Qwen-VL-7B) and SignReasoner (Pangu-MM-7B) and report the performance under the same benchmarks. We provide the post-training details of different VLMs and the Compute Reporting Form in our supplements.

\textbf{Evaluations.} 
For the TrafficSignEval benchmark, we design two distinct evaluation protocols respectively for manual and automatic evaluations: 
(1) TrafficSignEval-\textit{Caption Protocol}. 
Models are directly prompted to output a dense caption describing and reasoning the traffic sign contents. A strict human judgment is used to mark the correct answer only if all elements of the traffic sign are accurately described and the functions of sign are presented. 
(2) TrafficSignEval-\textit{Structure Protocol}.
Models are prompted to generate the FSU decomposition. Given the highly structured nature of the output, we designed an automated evaluation script that sequentially assesses all key-value pairs. For the open-set values, the prediction is marked correct only if its similarity with the GT exceeds a threshold of 0.8. Besides, the answer is marked as incorrect if it is unparsable. 
Details about our automatic evaluation can be found in our supplementary.

\begin{table*}[t]
    \centering
    \renewcommand\arraystretch{1.05}
    \setlength{\tabcolsep}{0.15mm}{
    \begin{tabular}{l|cccc|c|cccc|c}
    \hline
    \multirow{2}{*}{Model}   & \multicolumn{5}{c|}{TrafficSignEval-Caption} & \multicolumn{5}{c}{TrafficSignEval-Structure} \\\cline{2-11}
     & Direction  & Notice  & Lane & Const.  & Avg. & Direction  & Notice  & Lane & Const.  & Avg.   \\\hline
    \multicolumn{11}{l}{\textit{Zero-Shot Performance}}\\\hline 
    InternVL3-8B & 55.88 & 57.14& 53.40 & 50.00 & 54.07 & 51.43 & 63.64 & 54.84 & 57.14 & 55.38   \\
    InternVL3-78B  & 55.88 & 61.90 & 74.76 & 57.14 & \textbf{68.02} & 68.57 & 63.64 & 61.29 & 71.43 & \textbf{63.59} \\
    Qwen2.5-VL-7B & 47.06 & 66.67 & 42.72 & 64.29 & 48.26 & 28.57 & 45.45  & 36.29 & 64.29 & 37.95 \\
    Qwen2.5-VL-72B  & 64.71 & 71.43 & 63.11 & 71.43 & 65.12 & 40.00 & 68.18 & 50.81 & 71.43 & 52.31\\
    Pangu-MM-7B & 55.88 & 52.38 & 53.40 & 57.14 & 54.07 & 41.18 & 33.33 & 46.60& 35.71 & 43.02\\\hline
    \multicolumn{11}{l}{\textit{Caption SFT Performance}}\\\hline
    Qwen2.5-VL-7B & 58.82&61.90&56.31&64.29&58.14$(+9.88)$&28.57 & 45.45  & 36.29 & 64.29 & 37.95\\
    Pangu-MM-7B & 70.59&61.90&71.84&78.57&\textbf{70.93}$(+16.86)$&41.18 & 33.33 & 46.60& 35.71 & \textbf{43.02}\\\hline
     \multicolumn{11}{l}{\textit{Caption-FSU Distillation Performance
     }}\\\hline
    SignReasoner (Qwen2.5-VL-7B) & 70.59 & 66.67 & 76.70 & 57.14 & 72.67 $(+24.41)$ & 74.19 & 66.67 & 85.50 & 94.44&83.08 $(+45.13)$ \\
    SignReasoner (Pangu-MM-7B) & 79.41 & 66.67 & 83.50 & 64.29 & \textbf{79.07} $(+25.00)$ & 83.87 & 85.71 & 83.21 & 84.21&\textbf{83.59} $(+40.57)$ \\\hline
    \multicolumn{11}{l}{\textit{FSU-GRPO Training Performance}}\\\hline
    SignReasoner (Qwen2.5-VL-7B) & 71.00 & 67.31 & 76.90 & 57.67 & 73.24 $(+24.98)$ & 74.19 & 73.33 & 86.26 & 94.44&84.10 $(+46.15)$ \\
    SignReasoner (Pangu-MM-7B) & 80.01 & 67.76 & 83.98 & 64.90 & \textbf{79.98} $(+25.91)$ & 80.65
     & 80.00 & 84.73
      & 88.89 & \textbf{84.10} $(+41.08)$ \\\hline
    \end{tabular}}
    \caption{Comparing caption and structure protocols under TrafficSignEval.}
    \label{tab1}
\end{table*}

\subsection{Zero-Shot Prompting Baseline}
Existing VLMs claim superior performance in visual recognition, counting, and relational reasoning, etc. However, these results are typically derived from tests conducted in general domains. To the best of our knowledge, a dedicated capability analysis focused specifically on traffic sign comprehension has yet to be performed. Therefore, we systematically test different VLMs under TrafficSignEval-Caption and Structure Protocols as baselines. The results for zero-shot prompting are shown in Tab.\ref{tab1}.

The evaluation across both caption and structure protocols reveals critical limitations in current Vision-Language Models (VLMs). In the Caption Protocol, performance showed a positive correlation with model size (e.g., InternVL3-78B and Qwen2.5-VL-72B surpassing InternVL3-8B and Qwen2.5-VL-7B), but even the top model, InternVL3-78B, achieved only 68.0\% accuracy, falling far short of practical deployment needs. The Structure Protocol proved significantly more challenging due to output formatting, yielding invariably lower performance.

\subsection{Comparison with Post-Training Methods}

In this section, we contrast two post-training strategies.
The first is simply SFT the model with annotated captions, noted as ``Caption SFT" in the table. We manually annotated 5K traffic sign captions for training.
The second strategy is SignReasoner, that bases on FSU decomposition for Cap-FSU Iterative Distillation and FSU-GRPO training. 
Due to a higher complexity of FSU annotation, we sample 726 diverse images from the 5K traffic signs, covering the four FSU categories for training. 
The final comparative results are presented in the Tab.\ref{tab1}.
Several critical conclusions can be derived from the experimental results:

\textbf{Limitations of Caption SFT.} We first observe that the standard Caption SFT method yields a noticeable performance improvement over the zero-shot baselines of Qwen (+9.88\%) and Pangu (+16.86\%), with Pangu demonstrating a more significant gains because the annotated captions are initially generated by Pangu model and then through manual inspection. 
However, we found that training exclusively on caption data does not yield any gains for the Structure Protocol evaluation; instead, it often results in a performance decline.

\textbf{Synergistic Effect of Cap-FSU Distillation.} The Cap-FSU Iterative Distillation in the SignReasoner pipeline brings substantial simultaneous improvement under both the Caption (+25.00\%) and Structure (+40.57\%) evaluation protocols. Notably, the gain in Caption performance is significantly greater than that achieved by the direct Caption SFT method (+16.86\%), which corroborates our hypothesis regarding the synergistic enhancement between the captioning and FSU-Reasoning.

\begin{figure*}
	\centering
	\includegraphics[scale=0.64]{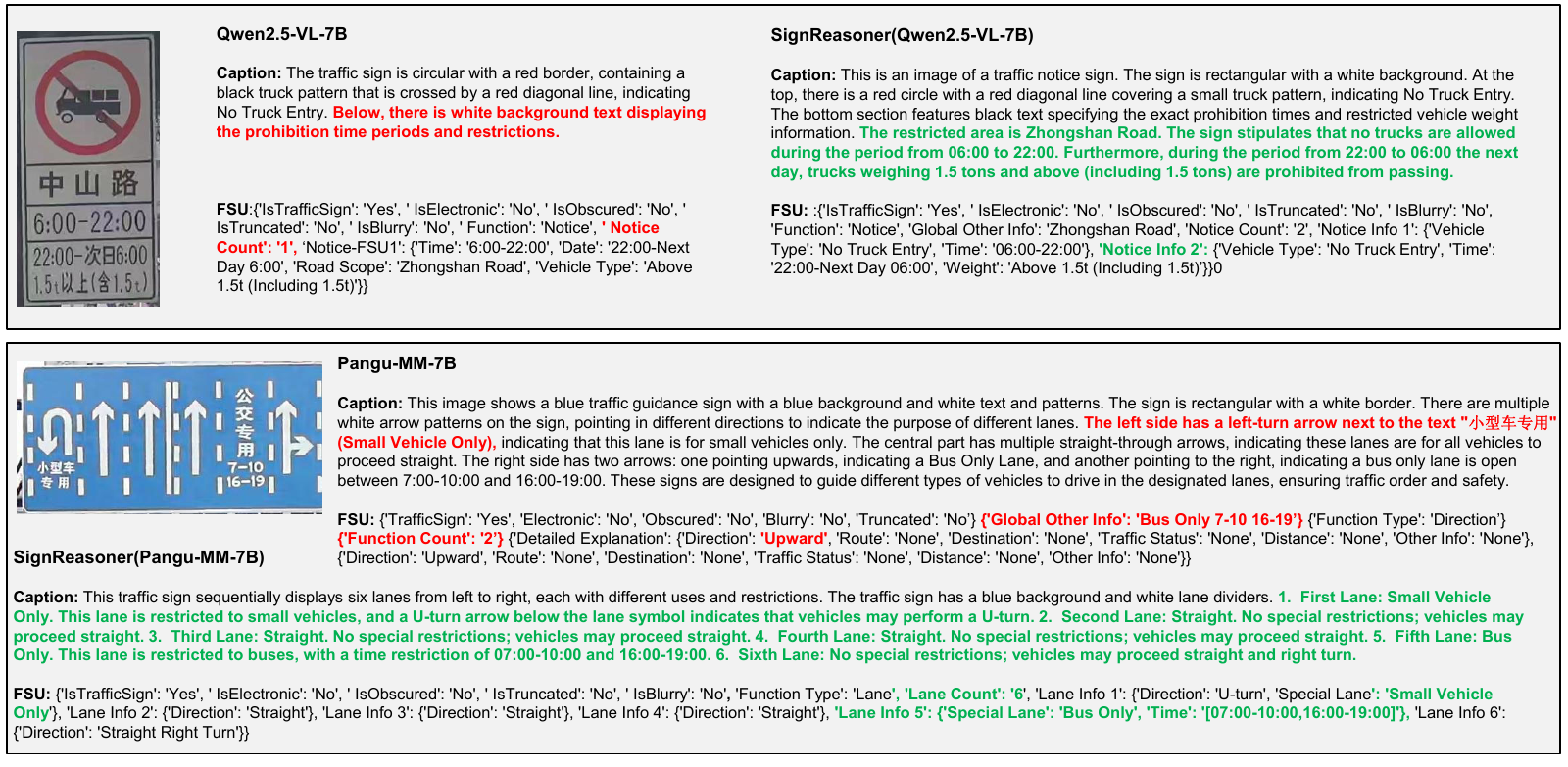}
	\caption{\textbf{Visual comparison between SignReasoner and base VLMs.}}
	\label{fig4}
\end{figure*}

\textbf{Effectiveness of FSU-GRPO Training.} The FSU-GRPO training of the SignReasoner pipeline improves both Qwen (+46.15\%) and Pangu (+41.08\%) on hard samples as visualized in Fig.\ref{fig4}. 
Importantly, the TED-reward that encourages a more accurate FSU decomposition brings improvements to the captioning, which may confirm that GRPO successfully boosts reasoning in captions.

\textbf{Data Efficiency and Superiority.} Finally, it is worth to note that, in contrast to simply using annotated captions for SFT, the SignReasoner training pipeline achieves a significant enhancement in both the Caption and Structure evaluation dimensions while utilizing only 726 annotated FSU samples. This demonstrates both the effectiveness and the data-efficiency of SignReasoner solution.

\begin{table}[t]
	\centering
	\renewcommand\arraystretch{1.05}
	\setlength{\tabcolsep}{0.5mm}{
		\begin{tabular}{ccc|cc}
			\hline
			\multicolumn{3}{c|}{FSU Design} & \multicolumn{2}{c}{SignReasoner}\\\hline
			Hierar. & Open & Closed& Qwen2.5-VL-7B & Pangu-MM-7B  \\\hline
			 &$\checkmark$&$\checkmark$& 80.27 & 79.48 \\
			 $\checkmark$&$\checkmark$&& 82.89 & 82.14 \\
			 $\checkmark$&$\checkmark$&$\checkmark$& \textbf{83.59} & \textbf{83.08} \\\hline
	\end{tabular}}
	\caption{Ablation on FSU Design.}
	\label{tab3}
\end{table}

\begin{table}[t]
	\centering
	\renewcommand\arraystretch{1.05}
	\setlength{\tabcolsep}{0.5mm}{
		\begin{tabular}{cc|cc}
			\hline
			{Caption-FSU} & {Iterative} & \multicolumn{2}{c}{SignReasoner}\\\cline{3-4}
			Format & Step  & Qwen2.5-VL-7B& Pangu-MM-7B  \\\hline
			&0& 74.87 & 78.26  \\
			$\checkmark$&1& 77.95 & 81.03 \\
			$\checkmark$&2& \textbf{83.08} & \textbf{83.59}  \\
			$\checkmark$&3& 82.37 & 83.06  \\\hline
			
	\end{tabular}}
	\caption{Ablation on Caption-FSU Iterative Distillation.}
	\label{tab4}
\end{table}

\begin{table}[t]
	\centering
	\renewcommand\arraystretch{1.05}
	\setlength{\tabcolsep}{0.5mm}{
		\begin{tabular}{ccc|cc}
			\hline
			\multicolumn{3}{c|}{Mixed Rewards} & \multicolumn{2}{c}{SignReasoner}\\\hline
			$R_{\text{C-FSU}}$ & $R_{\text{FSU}}$ & $R_{\text{TED}}$ & Qwen2.5-VL-7B& Pangu-MM-7B  \\\hline
			$\checkmark$&& & 83.14 & 83.67 \\
			&$\checkmark$& & 83.08 & 83.59\\\hline
			$\checkmark$& & $\checkmark$& 83.65 & 83.82 \\
			&$\checkmark$& $\checkmark$& 83.96 & 84.02 \\
			$\checkmark$&$\checkmark$& $\checkmark$& \textbf{84.10} & \textbf{84.10} \\\hline
	\end{tabular}}
	\caption{Ablation on Mixed Rewards Design.}
	\label{tab5}
\end{table}

\subsection{Ablation Study}
\label{sect4.5}
In this section, we conduct ablation study to verify the effectiveness of designs adopted in SignReasoner. The reported results adopt structure protocol for comparison.

\textbf{FSU Design.}
Firstly, we validate FSU configurations, including the hierarchical key-value and the hybrid value semantics. The results are reported in Tab.\ref{tab3}. The hierarchical key-value design brings about 3.32\% and 3.60\% gains respectively for Pangu and Qwen. Besides, using a pure open-set value semantic results in performance decline.

\textbf{Caption-FSU Iterative Distillation.}
Next, we investigate the Synergistic Effect brought about by the ``caption-FSU" output format and determine the optimal iteration steps for the caption distillation. The results are presented in Tab.\ref{tab4}. As can be seen, the ``caption-FSU" brings considerable improvements for both Pangu (3.12\%) and Qwen (2.77\%). Furthermore, the iterative distillation algorithm reach the optimal performance at the second iteration step.

\textbf{Mixed Rewards Design.}
Finally, we validated the improvements brought about by various reward functions. Experimental results are reported in Tab.\ref{tab5}. We find that the single format reward ($R_{\text{C-FSU}}$ or $R_{\text{FSU}}$) brings minor improvements. Combining the TED reward ($R_{\text{TED}}$) with other rewards brings the larger performance gains.

\subsection{Discussions of Limitations and Society Impact}
The current SignReasoner still has some limitations, including its reliance on cropped images (rather than full-scene understanding) and its restriction to Chinese traffic signs. However, by defining a formal, parsable FSU output, this work offers a standardized representation for VLM's traffic sign understanding which may hold promise for future Autonomous Driving VLMs, potentially leading to safer, more reliable, and more explainable driving decisions.

\section{Conclusion}
We propose SignReasoner to transform general VLMs into traffic sign experts, specifically addressing the lack of compositional generalization. Our innovations including the Functional Structure Unit (FSU), which enables function-based decomposition to learn the underlying grammar of complex signs and a two-stage VLM post-training pipeline involving Iterative Caption-FSU Distillation and FSU-GRPO , to boost the complex traffic sign reasoning. Experiments on TrafficSignEval show that SignReasoner reaches SOTA with remarkable data efficiency, improving sign understanding without any architectural modification.

{
    \small
    \bibliographystyle{ieeenat_fullname}
    \bibliography{main}
}

\clearpage
\setcounter{page}{1}
\maketitlesupplementary

\section{SignReasoner Implementations}
We elaborate on several implementation details of SignReasoner in this section, including the uniquely designed keys for each FSU (Sect.\ref{sect61}), the specific prompt and response format employed to instruct the VLM for both Captioning and FSU-Reasoning (Sect.\ref{sect62}), and the algorithmic implementation of Tree Edited Distance (Sect.\ref{sect63}) .

\subsection{Keys in FSUs}
\label{sect61}
We present the reserved keys for the four Functional Semantic Unit (FSU) categories in Tab. \ref{tabsupp1}.

\begin{table}[h]
	\centering
	\caption{Summary of possible keys in each kind of FSU.}
	\begin{tabular}{p{2cm}|p{5cm}}
		\hline
		\textbf{Functions} & \textbf{Possible Keys} \\
		\hline
		\textbf{Lane} & \texttt{Turn, Location, Special Lane, Time, Date, Speed, Weight, Height, Other Information} \\
		\hline
		\textbf{Direction} & \texttt{Direction, Via, Destination, Traffic Status, Distance, Other Information} \\
		\hline
		\textbf{Construction} & \texttt{Construction Site, Detour Information, Other Information} \\
		\hline
		\textbf{Notice} & \texttt{Direction, License Plate, Vehicle Type, Time, Date, Road Range, Speed, Weight, Height, Other Information}
		\\\hline
	\end{tabular}
	\label{tabsupp1}
\end{table}

\subsection{Prompts and Responses}
\label{sect62}
Three kinds of Prompts are adopted in our training data, including Captioning prompt $P_\text{{caption}}$, FSU-Reasoning prompt $P_\text{{reason}}$ and the Caption-FSU prompt $P_\text{cap-FSU}$, with each corresponding to a unique response format.

\textbf{Captioning.} The captioning prompt ($P_{\text{caption}}$) is designed to instruct VLMs to execute a comprehensive analysis, which entails vividly describing the visual content within traffic sign images, inferring the underlying relations among the sign's various elements, and ultimately articulating the traffic sign's significance and providing actionable driver guidance.

\textbf{FSU-Reasoning.}
In contrast, the prompt for FSU-Reasoning ($P_{\text{reason}}$) is primarily structured by providing the model with the predefined FSU categories and corresponding keys, subsequently instructing the model to perform the task of FSU decomposition. 

\textbf{Caption-FSU.}
Finally, the combined caption-FSU prompt ($P_{\text{cap-FSU}}$) is formulated by integrating the two prior prompt types. The model initially generates a descriptive caption following the instruction of the captioning prompt, and subsequently leverages this intermediate output to guide the FSU-Reasoning task.

The details of these three prompts followed by representative examples of the VLM's generated responses are presented in Tab.\ref{tabsupp2} and Tab.\ref{tabsupp3}.

\begin{table*}[t] 
	\centering
	\caption{Example Prompts and Corresponding Model Responses for SignReasoner.}
	\label{tabsupp2}
	
	\begin{tabularx}{\textwidth}{@{} c X @{}} 
		\toprule
		\textbf{ID} & \textbf{Prompt ($P$) / Model Responses} \\
		\midrule
		
		\textbf{Captioning} & $\bm P_{\textbf{caption}}$: 
		Please analyze this cropped traffic sign image in detail and generate a detailed caption describing the colors, shapes, and patterns of the signs, and explain their meanings. This part of the answer should be placed between the \textless\text{caption}\textgreater \ \ \textless/\text{caption}\textgreater \ tags.
		
		\textbf{Responses:} \textless\text{caption}\textgreater This traffic sign is blue in color and has the standard shape of a traffic direction sign, with a white crossroad pattern in the center. The sign displays the names and directions of three roads:
		1. Fulong Road (Fulong Rd) is located at the top of the sign, indicating a straight-ahead direction.
		2. Mingle Road (Mingle Rd) is on the left side of the sign, indicating a left turn.
		3. Yangtaishan Road (Yangtaishan Rd) is on the right side of the sign, indicating a right turn.
		The color and shape of the traffic sign are designed to be simple and clear, using the white crossroad pattern to clearly indicate the options for going straight ahead or turning left or right. The road names are written in white font, creating a sharp contrast with the blue background, making the information easy to recognize and read. The primary purpose of the traffic sign is to help drivers choose the correct road direction, ensuring safe and efficient travel.\textless/\text{caption}\textgreater \\
		\midrule
		
		\textbf{FSU-Reasoning} & $\bm P_{\textbf{Cap-FSU}}$: Please provide the information on the traffic sign in a structured format as follows:
		1. Provide the global attribute information of the traffic sign. The output key values include \{Traffic Sign, Electronic Sign, Obstruction, Blur, Truncation\}, with answers chosen from \{Yes, No\}. Provide the \{Other Global Information\} of the traffic sign, with answers selected from the traffic sign itself.
		2. Provide the functional category of the traffic sign. The answer should be chosen from \{Lane, Direction, Notice, Construction\}.
		3. Provide the number of functional information items on the sign (for lane information, this refers to the number of lanes).
		4. List detailed explanations for each functional information item. For lane information, the output key values include \{Electronic Sign, Turn, Location, Special Lane, Time, Date, Speed, Weight, Height, Other Information\}. For direction information, the output key values include \{Direction, Route, Destination, Traffic Status, Distance, Other Information\}. For construction information, the output key values include \{Construction Site, Detour Information, Other Information\}. For notice information, the output key values include \{Direction, License Plate, Vehicle Type, Time, Date, Road Range, Speed, Weight, Height, Other Information\}. Only provide answers based on the text and symbols on the traffic sign, without adding any extra explanatory text. Place the structured response within the \textless\text{FSU}\textgreater \ \ \textless/\text{FSU}\textgreater \  tags.
		
		\textbf{Responses:} \textless\text{FSU}\textgreater\{``Traffic Sign": ``Yes", ``Electronic Sign": ``No", ``Obstruction": ``Yes", ``Truncation": ``No", ``Blurriness": ``No", ``Function Type": ``Direction", ``Number of Direction Information": ``2", ``Direction Information 1": \{``Direction": ``Go Straight", ``Via": ``Li Yang Road", ``Destination": `` [The Bund, Haining Road] "\}, ``Direction Information 2": \{``Direction": ``Turn Right", ``Via": ``Li Yang Road", ``Destination": ``Obstruction"\}\}\textless/\text{FSU}\textgreater
		\\\midrule
	\end{tabularx}
\end{table*}\begin{table*}[t] 
\centering
\caption{Example Prompts and Corresponding Model Responses for SignReasoner.}
\label{tabsupp3}

\begin{tabularx}{\textwidth}{@{} c X @{}} 
	\toprule
	\textbf{ID} & \textbf{Prompt ($P$) / Model Responses} \\
	\midrule
	\textbf{Caption-FSU}  & $\bm P_{\textbf{FSU-Reason}}$: Please provide the information on the traffic sign in a structured format as follows:
	1. Please analyze this cropped image of the traffic sign in detail and generate a detailed caption describing the colors, shapes, and patterns of the signs, and explain their meanings. This part of the answer should be placed between the \textless\text{caption}\textgreater \ \ \textless/\text{caption}\textgreater \ tags.
	2. Based on the caption description, provide the global attribute information of the traffic sign. The output key values include \{Traffic Sign, Electronic Sign, Obstruction, Blur, Truncation\}, with answers chosen from \{Yes, No\}. Also, provide the \{Other Global Information\} of the traffic sign, with the answer selected from the traffic sign options.
	3. Based on the caption description, determine the functional category of the traffic sign. The answer should be chosen from \{Lane, Direction, Notice, Construction\}.
	4. Based on the caption description, indicate the number of functional information items on the sign (for lane information, this refers to the number of lanes).
	5. Based on the caption description, list detailed explanations for each functional information item. For lane information, the output key values include \{Electronic Sign, Turn, Location, Special Lane, Time, Date, Speed, Weight, Height, Other Information\}. For direction information, the output key values include \{Direction, Route, Destination, Traffic Status, Distance, Other Information\}. For construction information, the output key values include \{Construction Site, Detour Information, Other Information\}. For notice information, the output key values include \{Direction, License Plate, Vehicle Type, Time, Date, Road Range, Speed, Weight, Height, Other Information\}. Only provide the corresponding answers based on the text and symbols on the traffic sign, without adding any extra explanatory text. The structured answers for sections 2-5 should be placed between the \textless\text{FSU}\textgreater \ \ \textless/\text{FSU}\textgreater \ tags.
	
	\textbf{Responses:} \textless\text{caption}\textgreater This traffic sign is blue in color and has a standard traffic sign shape with a white crossroad pattern in the center. The sign displays the names and directions of three roads: 1. Fulong Road (Fulong Rd) is located at the top of the sign, indicating a straight ahead direction. 2. Mingle Road (Mingle Rd) is on the left side of the sign, indicating a left turn. 3. Yangtaishan Road (Yangtaishan Rd) is on the right side of the sign, indicating a right turn. The color and shape of the traffic sign are designed to be simple and clear, using the white crossroad pattern to clearly indicate the options for going straight ahead or turning left or right. The road names are written in white font, creating a sharp contrast with the blue background, making the information easy to recognize and read. The primary purpose of the traffic sign is to help drivers choose the correct road direction, ensuring safe and efficient driving.\textless/\text{caption}\textgreater\textless\text{FSU}\textgreater{``Traffic Sign": ``Yes", ``Electronic Sign": ``No", ``Blocked": ``No", ``Truncated": ``No", ``Blurred": ``No", ``Function Type": ``Direction", ``Number of Direction Information": ``3", ``Direction Information 1": {``Direction": ``Straight Ahead", ``Destination": `` [Fulong Road, Fulong Rd] "}, ``Direction Information 2": {``Direction": ``Left Turn", ``Destination": `` [Mingle Road, Mingle Rd] "}, ``Direction Information 3": {``Direction": ``Right Turn", ``Destination": `` [Yangtaishan Road, Yangtaishan Rd] "}}\textless/\text{FSU}\textgreater \\
	\bottomrule
\end{tabularx}
\end{table*}

\subsection{Tree Edited Distance}
\label{sect63}
This algorithm calculates the Tree Edit Distance (TED), which determines the minimum cost required to transform one tree structure into another using a set of editing operations (insertions, deletions, and modifications). We adopt this algorithm for measuring differences between predicted and ground truth FSU trees. Formally, the detail implementations are illustrated in alg.\ref{alg:tree_edit_distance}. It operates recursively in a bottom-up fashion. For internal nodes, the primary challenge is matching their unordered children. This is solved by constructing a cost matrix based on the recursive distances between all pairs of children and finding the optimal assignment (minimum weight matching) using the Linear Sum Assignment method (i.e., Hungarian Algorithm). For the ordered TED calculation, the assignment process is canceled.  The total distance is the sum of the current node's modification cost, the minimum cost of the optimal child assignment, and the costs for handling any unmatched (inserted or deleted) subtrees. 

\section{Training and Evaluation}

This section details the specifics of the training and evaluation datasets (Sect.\ref{sect71}), the training particulars (Sect.\ref{sect72})—including data construction, hyper-parameters, and hardware—and the comprehensive evaluation methodology on the Structure protocols (Sect.\ref{sect73}).
\subsection{Datasets}
\label{sect71}
\textbf{Training Data.}
We primarily utilize two types of training data. The first type is the data designated for Caption Supervised Fine-Tuning (SFT). This dataset consists of 5K clear traffic sign images acquired from on-board cameras. We employ the Pangu-MM-38B model for initial annotation, obtaining raw captions which are then subjected to manual quality assurance. The process involved correcting errors in the raw captions and augmenting them with the traffic sign's significance and corresponding driver guidance. 
The second data type is curated by selecting 726 images from the initial 5K set, ensuring maximum data diversity. This subset was then subjected to a more granular annotation process using the FSU decomposition. 
Notably, for the first type of data, we mainly adopt it for caption-based SFT, thus all the 5K data are organized in the Captioning prompt formats for training. For the second type of data, due to the synergistic effect as noted in Sect.\ref{sect3.2}, we use both the FSU-Reasoning and Caption-FSU formats to generate the training data, resulting in total of $1452$ ($i.e.,726\times 2$) samples.

\textbf{TrafficSignEval Benchmark.}
For evaluation purposes, we establish the first FSU-Reasoning benchmark TrafficSignEval, which comprises 195 distinct traffic sign images, all mutually exclusive from the training data, and is designed to reflect the VLM's understanding performance across each category.
Specifically, the benchmark includes 34 Direction signs, 21 Notice signs, 50 Lane signs, and 14 Construction signs. Each traffic sign within the benchmark is also accompanied by the FSU decomposition annotation to facilitate subsequent automated evaluation.
All the training and evaluation data are traffic signs cropped from the driving scene images, with the average resolution of $265\times225$.

\begin{figure*}[h]
	\centering
	\includegraphics[scale=0.6]{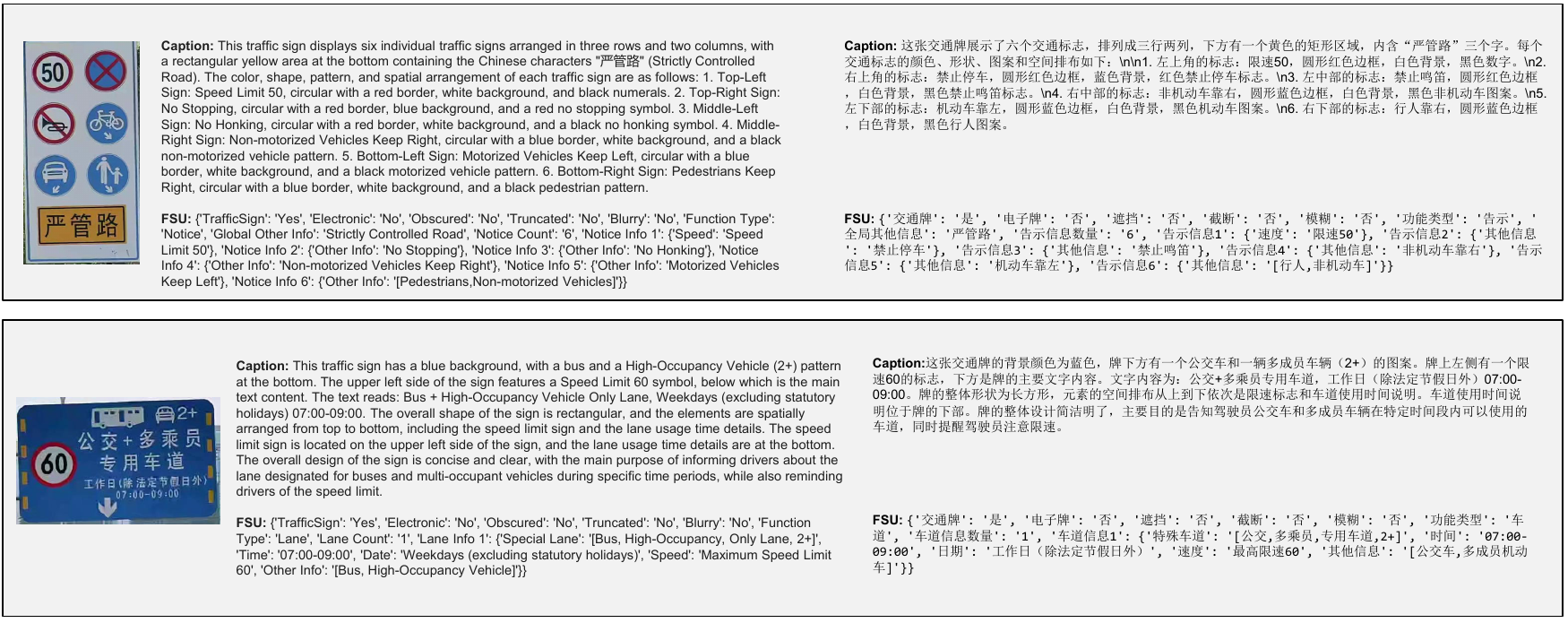}
	\caption{\textbf{Visualizations of Traffic Sign Understanding in SignReasoner (Pangu-MM-7B).}}
	\label{figsupp2}
\end{figure*}

\begin{figure*}[h]
	\centering
	\includegraphics[scale=0.6]{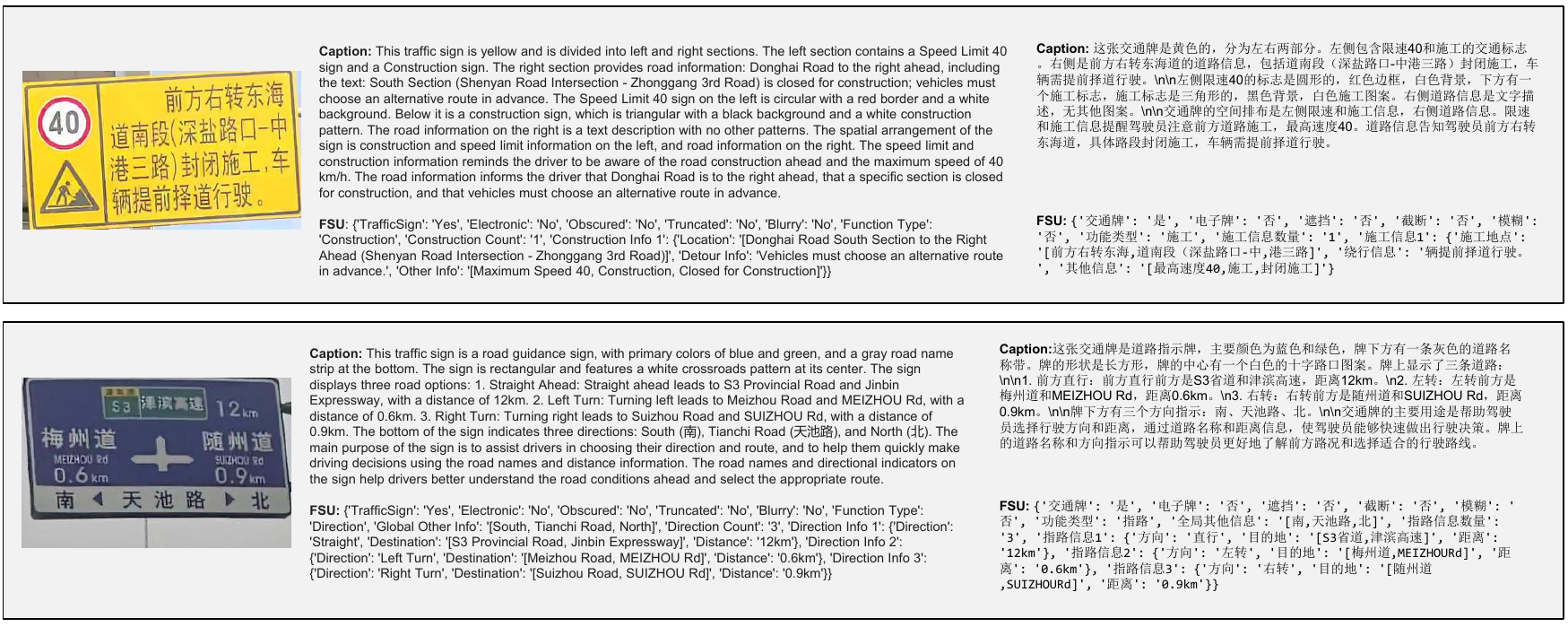}
	\caption{\textbf{Visualizations of Traffic Sign Understanding in SignReasoner (Pangu-MM-7B).}}
	\label{figsupp3}
\end{figure*}

\subsection{Training Details}
\label{sect72}
There are mainly two training stages in SignReasoner, the Caption-FSU Distillation and the FSU-GRPO.

In the first stage, we SFT the full parameters of base VLM (i.e., Qwen2.5-VL-7B and Pangu-MM-7B) via the Caption-FSU and FSU-Reasoning formatted SFT data. Both VLMs are trained 20 epochs with the learning rates set to $2\times10^{-5}$. The min learning rate is set to $2\times10^{-6}$ under cosine decay. The training is implemented on eight 8-GPU nodes with global batch size set to 64. The max encoder sequence length is set to 4096. We follow the ablation results to set the iterative steps as $2$ for the best performance. 

For the second stage, we adopt the GRPO algorithm for RL training on VLM's full parameters via the Caption-FSU formatted data on eight 8-GPU nodes. The GRPO training uses 1000 iters with learning rate set to $1\times10^{-6}$ and global batch size set to 32. The roll-out number is set to 8, with the roll-out batch size set to 32. For the GRPO hyperparameters, we set gamma to 1.0, lambda to 0.95 and KL coefficient to 0.05.
All three rewards are adopted to form the mixed rewards $R_{\text{Mixed}}$ as in Eq.\ref{eq6}. The TED algorithm is illustrated in Sect. \ref{sect63}. We set $\sigma_1 = 0.5, \sigma_2 = 5, \sigma_3 = 0.5$ to stably transform the TED reward into the range [0,1]. The max encoder sequence length is set to 4096.

All the training is implemented under the Ascend 910B3 GPU with PyTorch framework. Details about the GPU have been illustrated in the Compute Reporting Form.

\subsection{Evaluation Details}
\label{sect73}
\begin{figure}[h]
	\centering
	\includegraphics[scale=0.64]{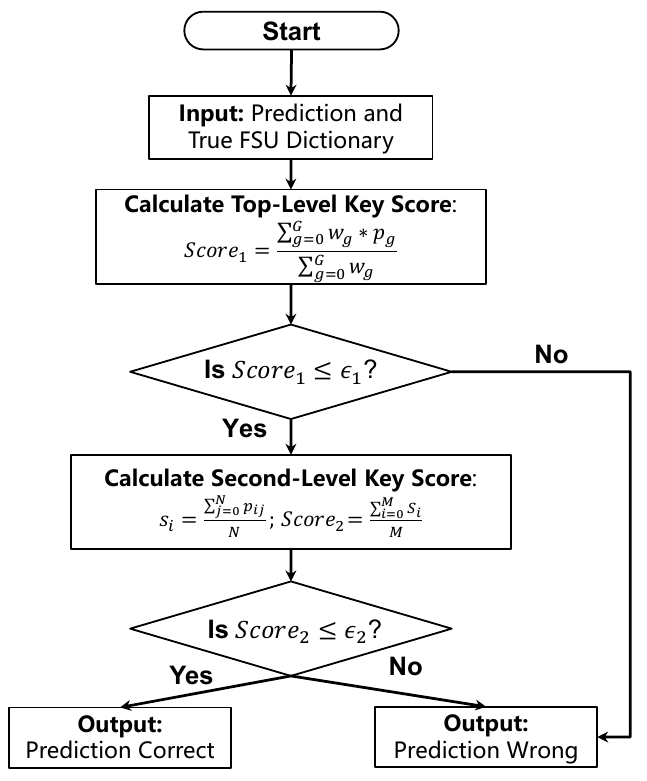}
	\caption{\textbf{Flowchart of automatic evaluation algorithm.}}
	\label{figsupp1}
\end{figure}

The specific process of automated evaluation in the TrafficSignEval-Structure protocol is illustrated in Fig.\ref{figsupp1}. The input consists of the Predicted and True FSU Dictionaries. First, the matching score, $\text{Score}_1$, for the top-level keys is computed. This score is a weighted sum where $p_g$ is the binary match indicator for the $g$-th top-level key ($p_g=1$ for a match, and $p_g=0$ otherwise). For top-level keys, a match is counted only if their respective values are strictly identical. $w_g$ denotes the weight assigned to each key, with greater weight assigned to critical keys (e.g., \texttt{Function} and \texttt{FSU Count}, etc.). $\text{Score}_1$ is then compared against a threshold $\epsilon_1 (0.8)$: if the score falls below the threshold, the sample is classified as incorrect; otherwise, the evaluation proceeds to the Second-Level Key Score calculation. In this stage, the matching score $s_i$, for each FSU is firstly calculated. Here, $p_{ij}$ indicates whether the $j$-th key (out of a total $N$ keys) within the $i$-th FSU matches. String similarity (with a threshold of 0.5) is employed to determine the match, where $p_{ij}=1$ for a match and $p_{ij}=0$ otherwise. Finally, the scores ($s_i$) of all $M$ FSUs are averaged to yield the Second-Level Key $\text{Score}_2$. If $\text{Score}_2$ falls below the threshold $\epsilon_2 (0.5)$, the sample is judged as incorrect; otherwise, it is classified as a correct sample.

\section{More Visualizations}
\label{sect8}
In this section, we provide more comprehensive visualized results (Fig.\ref{figsupp2}-\ref{figsupp3}) in both Chinese and English to demonstrate the powerful capabilities of SignReasoner.

\begin{algorithm*}
	\caption{Tree Edit Distance}
	\label{alg:tree_edit_distance}
	\begin{algorithmic}[1]
		\REQUIRE Tree nodes $n_1$, $n_2$
		\ENSURE Minimum edit cost
		
		\STATE \textbf{Function} \textsc{Helper}($n_1, n_2$):
		\STATE $cost \leftarrow 0$
		
		\IF{\textsc{IsLeaf}($n_1$) \textbf{and} \textsc{IsLeaf}($n_2$)} \STATE{\textit{\# Case 1: Both are leaf nodes}}
		\IF{$n_1.name = n_2.name$}
		\IF{$n_1.value \neq n_2.value$}
		\STATE $cost \leftarrow cost + 1$ \hfill \textit{\# Same name, different value}
		\ENDIF
		\ELSE
		\STATE $cost \leftarrow cost + 2$ \hfill \textit{\# Different name: replace cost}
		\ENDIF
		\ELSIF{\textsc{IsLeaf}($n_1$)} \STATE{\textit{\# Case 2: Only $n_1$ is leaf}}
		\STATE $cost \leftarrow cost + n_2.size + 1$ \hfill \textit{\# Insert subtree $n_2$}
		\ELSIF{\textsc{IsLeaf}($n_2$)} \STATE{\textit{\# Case 3: Only $n_2$ is leaf}}
		\STATE $cost \leftarrow cost + n_1.size + 1$ \hfill \textit{\# Delete subtree $n_1$}
		\ELSE 
		\STATE \textit{\# Case 4: Both are internal nodes}
		\IF{$n_1.name \neq n_2.name$}
		\STATE $cost \leftarrow cost + 1$ \hfill \textit{\# Modify current node name}
		\ENDIF
		
		\STATE $C_1 \leftarrow n_1.children, \quad C_2 \leftarrow n_2.children$
		\STATE $M \leftarrow |C_1|, \quad N \leftarrow |C_2|$
		\STATE Initialize cost matrix $D$ of size $M \times N$
		
		\FOR{$i=0$ \textbf{to} $M-1$}
		\FOR{$j=0$ \textbf{to} $N-1$}
		\STATE $D_{i,j} \leftarrow \textsc{Helper}(C_1[i], C_2[j])$ \hfill \textit{\# Recursive calculation}
		\ENDFOR
		\ENDFOR
		
		\STATE $(min\_sum, pairs) \leftarrow \textsc{LinearSumAssignment}(D)$ \hfill \textit{\# Hungarian Algo}
		\STATE $cost \leftarrow cost + min\_sum$
		
		\STATE $rows \leftarrow \{r \mid (r, c) \in pairs\}$
		\STATE $cols \leftarrow \{c \mid (r, c) \in pairs\}$
		
		\FOR{$i=0$ \textbf{to} $M-1$}
		\IF{$i \notin rows$}
		\STATE $cost \leftarrow cost + C_1[i].size + 1$ \hfill \textit{\# Unmatched child: delete}
		\ENDIF
		\ENDFOR
		
		\FOR{$j=0$ \textbf{to} $N-1$}
		\IF{$j \notin cols$}
		\STATE $cost \leftarrow cost + C_2[j].size + 1$ \hfill \textit{\# Unmatched child: insert}
		\ENDIF
		\ENDFOR
		\ENDIF
		
		\STATE \textbf{return} $cost$
	\end{algorithmic}
\end{algorithm*}


\end{document}